%% file: paper.tex
\documentclass[letterpaper, 10 pt, conference]{ieeeconf}  

\IEEEoverridecommandlockouts                              

\usepackage{graphicx}
\usepackage{subcaption}
\usepackage{amsmath} 
\usepackage{amssymb}  
\usepackage{xspace}
\usepackage[pstricks1-10]{vaucanson-g}
\usepackage{xcolor}
\usepackage{array}
\usepackage{tikz}
\usepackage[free-standing-units=true]{siunitx}

\input{macros}

\title{\LARGE \bf
Iterator-Based Temporal Logic Task Planning}

\author{Sebasti\'an A. Zudaire$^{1}$,
Martin Garrett$^{2}$,
and Sebasti\'an Uchitel$^{3}$%
\thanks{$^{1}$Instituto Balseiro - Universidad Nacional de Cuyo, Argentina.
        {\tt\small sebastian.zudaire@ib.edu.ar}}%
 \thanks{$^{2}$CNEA, Argentina.
                {\tt\small medgarrett@cab.cnea.gov.ar}}%
\thanks{$^{3}$Universidad de Buenos Aires,
Buenos Aires, Argentina and Imperial College London, UK.
        {\tt\small suchitel@dc.uba.ar}}%
}

\begin{document}

\onecolumn
© 2020 IEEE. Personal use of this material is permitted. Permission from IEEE must be
obtained for all other uses, in any current or future media, including
reprinting/republishing this material for advertising or promotional purposes, creating 
new
collective works, for resale or redistribution to servers or lists, or reuse of any 
copyrighted
component of this work in other works.

\bigskip

Author-submitted article to: \textbf{International Conference on 
Robotics and Automation (ICRA) 2020}

\twocolumn

\maketitle
\thispagestyle{empty}
\pagestyle{empty}

\begin{abstract}
Temporal logic task planning for robotic systems suffers from state explosion when specifications involve large numbers of discrete locations. 
We provide a novel approach, particularly suited for tasks 
specifications with 
universally quantified locations, that has constant time with respect 
to the number of 
locations, enabling synthesis of plans for an 
arbitrary number of them. 
We propose a hybrid control framework that uses an iterator to 
manage the discretised workspace hiding it from a plan enacted by 
a discrete event controller. 
A downside of our approach is that it incurs in increased overhead when executing a 
synthesised plan. We demonstrate that the overhead is reasonable for missions of a 
fixed-wing Unmanned Aerial Vehicle in simulated and real scenarios for up to 
\num{700000} locations. 

\end{abstract}


\input{intro}

\begin{figure*}[bt]
\centering
  \vspace{1mm}
 \includegraphics[scale=0.4]{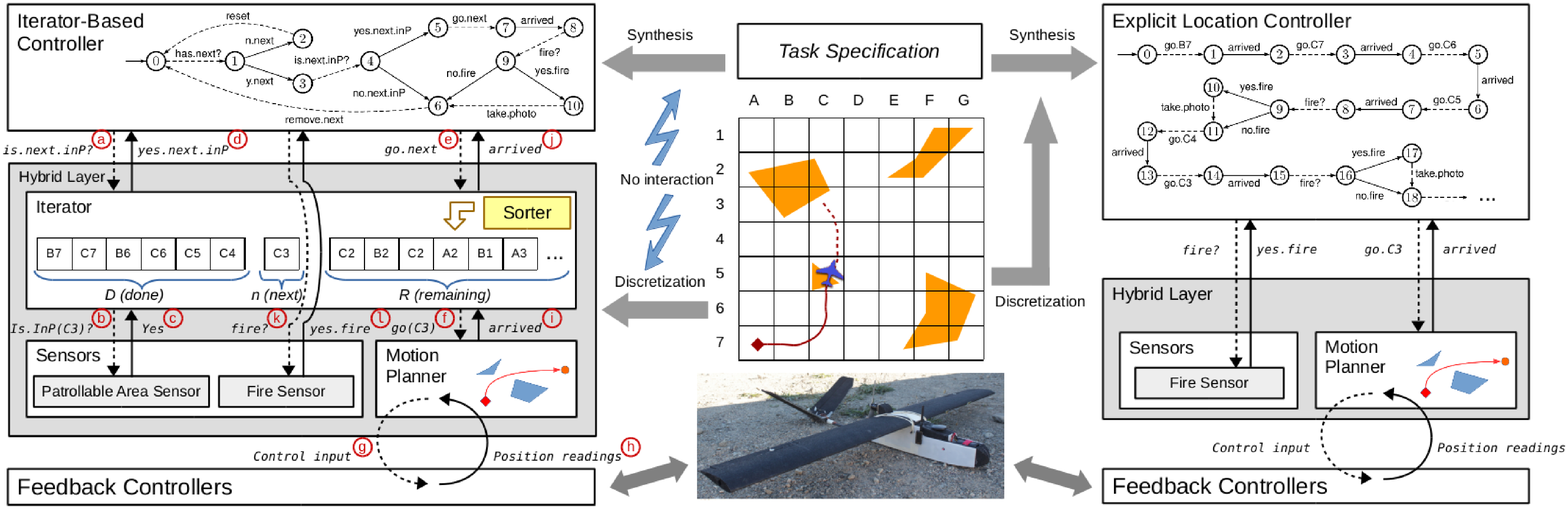}
 \caption{System Architecture for Iterator-Based and Explicit location plans. The 
fixed-wing UAV has a wingspan of \SI{1.6}{\metre}}
\label{fig:architecture}
\end{figure*}

\input{preliminaries}

\input{abstraction}
\input{hybrid}

\input{experiments}

\input{conclusions}

\IEEEtriggeratref{18}
\bibliographystyle{IEEEtran}
\bibliography{IEEEabrv,references}

\end{document}

%% file: macros.tex
\newcommand{\set}[1]{\{#1\}}
\newcommand{\gr}{{GR(1)}\xspace}

\newcommand{\G}{\square}
\newcommand{\F}{\lozenge}

\newcommand{\W}{\,\mathbf{W} \,}
\renewcommand{\implies}{\ensuremath{\Rightarrow}}

\newcommand{\comments}[1]{}

\newcommand{\caja}[1]{\raisebox{.5pt}{\textcircled{\raisebox{-.0pt} 
{{\scriptsize \textcolor{black}{#1}}}}}} 

\newcommand{\formatLoc}[1]{\formatAction{#1}}

\newcommand{\Processed}{\ensuremath{D}\xspace}

\newcommand{\Next}{\ensuremath{n}\xspace}
\newcommand{\Remaining}{\ensuremath{R}\xspace}

\newcommand{\formatAction}[1]{{\footnotesize \textsf{#1}}\xspace}

\newcommand{\photo}{\formatAction{take.photo}}
\newcommand{\fire}{\formatAction{fire?}}
\newcommand{\yFire}{\formatAction{yes.fire}}
\newcommand{\nFire}{\formatAction{no.fire}}
\newcommand{\inP}{\formatAction{is.next.inP?}}
\newcommand{\yinP}{\formatAction{yes.next.inP}}
\newcommand{\ninP}{\formatAction{no.next.inP}}
\newcommand{\go}{\formatAction{go.next}}

\newcommand{\hasNext}{\formatAction{has.next?}}

\newcommand{\remove}{\formatAction{remove.next}}
\newcommand{\reset}{\formatAction{reset}}
\newcommand{\yesNext}{\formatAction{y.next}}
\newcommand{\noNext}{\formatAction{n.next}}

\newcommand{\arrived}{\formatAction{arrived}}

\newcommand{\isPatrolNext}{\formatAction{is.next.inP?}}
\newcommand{\yesPatrolNext}{\formatAction{yes.next.inP}}

\newcommand{\yesAdjacentNext}{\formatAction{y.adjacent.next}}

\newcommand{\formatFluent}[1]{{\small \emph{#1}}\xspace}

\newcommand{\FpatrolNextYes}{\formatFluent{MustPatrol}}

\newcommand{\Farrived}{\formatFluent{Arrived}}
\newcommand{\VisitCondition}{\formatFluent{VisitCondition}}
\newcommand{\FrespondedPatrol}{\formatFluent{PatrolAnwered}}
\newcommand{\ArrivedCondition}{\formatFluent{ArrivedCondition}}

\newcommand{\FPhoto}{\formatFluent{PhotoTaken}}
\newcommand{\FYesFire}{\formatFluent{FireDetected}}
\newcommand{\FrespondedFire}{\formatFluent{FireAnswered}}

\newboolean{showcomments}
\setboolean{showcomments}{true} 

\ifthenelse{\boolean{showcomments}}
{
	\newcommand{\del}[1]{\textcolor{red}{\sout{#1}}}    
}{
	\newcommand{\del}[1]{}                              
	
}

\ifthenelse{\boolean{showcomments}}{
	\newcommand{\nbc}[3]{
		{\colorbox{#3}{\bfseries\sffamily\scriptsize\textcolor{white}{#1}}}
		{\textcolor{#3}{\sf\small$\langle$\textit{#2}$\rangle$}}}
}{
	\newcommand{\nbc}[3]{}
	
}


%% file: intro.tex
\section{Introduction}

Discrete event controller synthesis is receiving increased attention as a means
for providing robot applications correct-by-construction task
plans (e.g., ~\cite{Fainekos05,Wolff13ICRA,Castro15}).
Synthesis from temporal logic specifications requires a discrete abstraction of
the environment to establish a discrete event model that can be analysed exhaustively
to produce task plans. 

Synthesis algorithms are computationally complex (e.g.,~\cite{piterman06} is
polynomial) with respect to the number of states of the discrete 
model. 
Hence, it
is crucial to establish an abstraction of the environment that is sufficiently fine 
grained
to allow appropriately capturing task requirements but coarse enough so as to not
making synthesis intractable.

A robot's workspace may be naturally discretised to the sensors' 
capabilities, e.g., land mapping with a low-autonomy Unmanned 
Aerial Vehicle (UAV) can require over \num{400} discrete locations~\cite{Mapping}.
The number of discrete locations can induce a combinatorial growth in the size of the 
discrete event model, which in turn can make synthesis intractable. 

We provide a novel approach that allows scaling the number of locations in task planning 
by exploiting the following observation: Many robot tasks specifications are, or 
can be, expressed as a universal quantification over a set of locations (e.g.,  
\textit{``For all locations in the discrete workspace, if the location satisfies 
... then visit it and do ... if ...''}). Examples 
include tasks 
in~\cite{Fainekos05,ReactiveMissionMotion,Wolff2013,Waldo,Livingston13, 
Nedunuri14,Vasile14} 
and~\cite{Menghi19}, where common robot task are surveyed. The common ground in 
these papers is the explicit management of locations, that makes synthesis intractable 
when increased. Indeed they do not report building plans for over \num{1200} 
locations.


Our approach uses a hybrid control framework \cite{Gazit08}, which can 
work with any motion planner \cite{Ji15}, in which synthesised task plans 
execute 
over an API that
provides an iterator that manages and hides the discretised workspace, offering the
plan one location at a time. Plans are synthesised from a specification that includes the
task requirements and a model of iterator that abstracts the number of locations that it
manages. Hence, the \textit{synthesis time is constant} with respect to these locations.

The price to be paid for constant synthesis time with respect to locations managed by the 
iterator is at runtime: Plans can only make decisions and act upon the location currently 
offered by the iterator and cannot refer to locations explicitly. That is, a plan cannot 
request going to a named location $x$, rather it must iterate over locations asking for 
each one if it is location $x$ (similar to the sensor-based approach 
in~\cite{ReactiveMissionMotion}). Thus, the order in which the iterator selects locations 
impacts the overall robot behaviour. A particularly bad case is if in a scenario with 
millions of locations, the iterator offers location $x$ at the very end.

We show that a hybrid control layer in which location sorting uses 
shortest trajectory is fast enough to provide acceptable though sub-optimal flight 
paths for tasks involving hundreds of thousands of locations (significantly beyond what 
synthesis with explicit location management is capable of) for a
fixed-wing UAV. The design is complementary to work on motion 
and trajectory planning~\cite{Bezzo16}. Indeed, more sophisticated reasoning below the 
discrete plan layer can be modularly 
included into the hybrid controller and could provide enhanced performance.

In summary, we present a hybrid controller approach aimed at tasks specifications with 
universally quantified locations that does not suffer from synthesis scalability 
limitations with respect to the number of locations. Task plans are synthesised from 
specifications given as Labelled Transition Systems and Fluent Linear Temporal Logic.  
Despite using simple location motion planning and trajectory control approaches, we 
demonstrate by simulating and flying four tasks: search and 
follow~\cite{ReactiveMissionMotion}, search and map~\cite{mavproxy}, 
patrol~\cite{Wolff2013}, and cover~\cite{Wei18}, that the approach can scale to hundreds 
of thousands of discrete locations.

%% file: preliminaries.tex
\section{Preliminaries}


\textbf{Labelled Transition System:} (LTS)~\cite{Keller76} are automata where
transitions are labelled with actions that constitute the interactions of the modelled 
system with its environment.
We partition actions into controlled and uncontrolled to specify assumptions about 
the environment and safety requirements for a controller. Figure~\ref{fig:sensor} 
models the assumption 
that \yFire and \nFire are responses to \fire, and the safety 
property that \fire is not issued before the response to a previous \fire. 

Complex models can be constructed by LTS composition. We use a standard definition of 
parallel composition ($\|$) that models the asynchronous execution of LTS, interleaving 
non-shared actions and forcing synchronisation of shared actions.

%

\textbf{Fluent Linear Temporal Logic:} (FLTL)~\cite{gianna03} is also used to describe
environment assumptions and task requirements. 
FLTL is a linear-time temporal logic that uses fluents to describe states over 
sequences of actions.  

A fluent \formatFluent{fl} is defined by a set of initiating actions, a set of 
terminating actions, and an initial value. We may omit set notation for singletons, 
e.g.,
$\formatFluent{Going} = \langle \go, \arrived \rangle_{\emph{Initially}_{\bot}}$
We may use an action label $\ell$ for the
fluent defined as $\formatFluent{fl} = \langle \ell,
\emph{Act}\setminus\set{\ell} \rangle$. Thus, the fluent \remove is 
only true just after the occurrence of the action \remove. 

FLTL is defined similarly to propositional LTL but where a fluent holds at a position $i$ in a 
trace $\pi$ based on the events occurring in $\pi$ up to $i$. Temporal connectives are 
interpreted as standard: $\F \varphi$, $\G \varphi$, and $ \varphi \W \psi$ mean that 
$\varphi$ eventually holds, always holds and holds until $\psi$ respectively. 

%



\textbf{Discrete Event Controller Synthesis} is defined as follows: Given an 
LTS  
$E$ with a set of controllable actions $L$, assumption $A$ and
 goal $G$ expressed in FLTL, find an
 LTS $C$ 
 such that $E\|C$) is deadlock free, $C$ does not block any
 non-controlled actions,
 and for every trace of $E\|C$ if the
 trace satisfies the assumption $A$, then the trace satisfies 
 $G$.


When goals and assumptions are restricted to a \gr form~\cite{piterman06} the control 
problem can solved in polynomial time. MTSA~\cite{MTSA} solves \gr control 
problems expressed with LTS and FLTL, requiring 
assumptions and 
goals in FLTL to be either \textit{i)} of the form $\bigwedge_{i=1}^n \G\F 
\varphi_i$ where $\varphi_i$ are Boolean combinations of fluents, or \textit{ii)} safety 
properties~\cite{Lamport77}.

%% file: abstraction.tex
\section{Discrete Abstraction} \label{sec:abstraction}

\begin{figure*}[bt]
\vspace{1mm}
\begin{subfigure}[t] {0.162\textwidth}
\input{Imagenes/LTS/iterator}
\caption{}
\label{fig:iterator}
\end{subfigure}
~
\begin{subfigure}[t]{0.092\textwidth}
\input{Imagenes/LTS/SensorLTS}
\caption{}
\label{fig:sensor}
\end{subfigure}
~
\begin{subfigure}[t] {0.117\textwidth}
\input{Imagenes/LTS/SensorNextLTS}
\caption{}
\label{fig:sensorNext}
\end{subfigure}
~
\begin{subfigure}[t] {0.117\textwidth}
\input{Imagenes/LTS/constraintNext}
\caption{}
\label{fig:constraintNext}
\end{subfigure}
~
\begin{subfigure}[t]{0.11\textwidth}
\input{Imagenes/LTS/constraintCurrent}
\caption{}
\label{fig:constraintCurrent}
\end{subfigure}
~
\begin{subfigure}[t]{0.10\textwidth}
\centering
\input{Imagenes/LTS/capabilities}
\caption{}
\label{fig:capabilities}
\end{subfigure}
~
\begin{subfigure}[t]{0.17\textwidth}
\centering
\input{Imagenes/LTS/ConstratintGo}
\caption{}
\label{fig:ConstraintGo}
\end{subfigure}

\vspace{-0.5mm}

\caption{Dashed (non-dashed) lines are controlled (uncontrolled) actions. (a) 
Iterator Model. (b) Fire sensor for current location. (c) 
Patrollable area sensor. (d) Constraint: \isPatrolNext only when \yesNext. (d) 
Constraint: \photo and \fire only when arrived. (e) Simplified UAV capabilities 
for Fire Patrol task. (f) Constraint: \go only when \yesNext.} 
\label{fig:models}
\end{figure*}
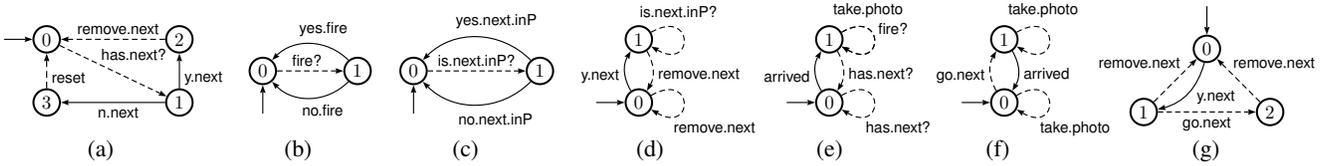

\subsection{Iterator-Based Task Plans} 
Consider a task for a UAV in which various locations of a grid-based map must be 
patrolled and photographs must be taken if fire is detected. A
plan for such a task in a temporal logic task and motion planning 
approach (e.g., ~\cite{Waldo}) might look like the LTS on the top right of 
Figure~\ref{fig:architecture} (\emph{Explicit Location Controller}) where the 
locations to be patrolled are $P = 
\set{\formatLoc{C5,A2,B2,\ldots}}$ (orange areas in Figure~\ref{fig:architecture}). 
The plan sequentially visits each location, checks for fire and takes a picture 
accordingly. Although the size of the plan grows linearly with $P$, 
the state space over which it is computed grows exponentially (i.e., $2^{\|P\|}$) as 
it must at least capture all possible orders in which locations in $P$ could be visited. 

A more compact plan for the same task may be synthesised if a richer 
execution 
environment is assumed. Consider an iterator that abstracts the size of discrete 
workspace and which can provide its locations, one at a time. In this case, a plan (top 
left 
of Figure~\ref{fig:architecture}) 
could consist of a loop iterating over the locations, checking for each location 
if it requires patrolling (\inP) and if so (\yinP) going to the location (\go) and 
upon 
arrival (\arrived) checking if the current UAV location has fire (\fire) and taking a 
photo (\photo) if needed. As locations are not explicitly treated in the plan, its size 
does not depend neither on the total number of locations nor the size of $P$. Similarly, 
the state space from which the plan can be synthesised is not affected, achieving 
\textit{constant synthesis time} with respect to the number of locations. 
In the remainder of this section we report on how to specify and synthesise 
iterator plans.

\subsection{Specification of Iterator-Based Task Plans}

Three aspects of the system must be abstracted to obtain a discrete event model 
from which to synthesise iterator-based task plans: the iterator, sensors 
and actuators. 

\subsubsection{Iterator Abstraction} 

The iterator is an abstract data type that manages a set of discrete locations 
$\mathcal{L}$ derived though the discretization of a region. We chose for simplicity to 
use grid-based maps as in~\cite{Wolff13ICRA,Livingston13,Wei18,Kundu19}.

An iterator is a triple $\langle \Processed, \Next, \Remaining \rangle$ where 
$\Processed$ and 
$\Remaining$ are sets of locations representing 
those that the plan has already processed ($D$one) and those that remain 
($R$emaining), and 
$\Next$ is the next location to be processed by the plan. The iterator is initialised as 
follows: $\Next$ is set to one element of $\mathcal{L}$, $\Remaining = \mathcal{L} 
\setminus \set{\Next}$, $\Processed = \emptyset$. We define the following 
operations:
\begin{itemize}
 \item \hasNext: returns true if and only if $\Next$ is not null.
 \item \remove: adds \Next to \Processed and if $\Remaining = \emptyset$  
 sets \Next to null, otherwise sets 
\Next to a location in \Remaining and \Remaining = \Remaining $\setminus \set{n}$.
 \item \reset: the iterator is reinitialised.
\end{itemize}

In Figure~\ref{fig:iterator} we depict an LTS that models interactions with an iterator. 
We model the return values of \hasNext with two different events \yesNext and 
\noNext. These two events are defined to be uncontrollable, i.e., it is the iterator and 
not the plan that decides whether the response is \yesNext or \noNext. 
A requirement such as $\G\F \hasNext \wedge 
\G (\yesNext \implies  (\neg \remove  \W  \go))$ will make 
the robot continuously visit all the locations with which the iterator was initialised.

\subsubsection{Sensors}
Similarly to~\cite{ReactiveMissionMotion}, we introduce binary sensors to 
model interaction with the environment. 
In this iterator-based setting, sensors can answer queries regarding the location that 
the 
robot is at and/or for the next location selected by the 
iterator. 
%
%
For example, Figure~\ref{fig:sensor} models a fire sensor that can be queried about 
the 
existence of fire at the current robot location. Figure~\ref{fig:sensorNext} shows an 
abstraction of the sensor that responds if the next location selected by the iterator is 
one that must be patrolled. 

For sensing over the next location, 
attribute $\Next$ must not be null. An additional LTS is included to 
constrain the occurrence of \isPatrolNext queries to between \yesNext and 
\remove 
as in Figure~\ref{fig:constraintNext}.

Additional constraints are typically needed for sensing over the current location to 
ensure that the plan is aware of what the current location is. 
For example, sensing for fire (\fire) and taking a photo (\photo)  
should occur between having \arrived to a particular 
location and starting to analyse the next (\hasNext),  
Figure~\ref{fig:constraintCurrent}.

\subsubsection{Primitive Capabilities}
Using a control-driven discretization~\cite{StateOfTheArt}, we 
define controllable/uncontrollable pairs to model the start/end of control 
modes~\cite{ControlModes}. For example, \go  
commands the robot to move to the next location according to the Iterator, and the 
uncontrollable action \arrived indicates that the target location has been reached. 
Other capabilities may be reasonably modelled as instantaneous 
such as $\photo$.

In Figure~\ref{fig:capabilities}, we depict the minimal capability model of a robot for 
the Fire Patrol task.
We also require that the \go command only be issued when the iterator has a next 
location to be processed (Figure~\ref{fig:ConstraintGo}).

\subsection{Task Specification Example} \label{sec:firetask}

To help understand how the abstraction described above works, we elaborate on
how the Fire Patrol task can be specified to obtain the plan 
shown in top left of Figure~\ref{fig:architecture}. 

We build an environment model $E$ as a parallel composition that describes 
assumptions and constraints on how the infrastructure on which the plan will execute. For 
the Fire Patrol task the composition includes exactly all the LTS described above: the 
iterator, the fire and patrol sensors with associated constraints and the robot 
capabilities and constraints, as depicted in Figure~\ref{fig:models}.
. 


We structure the task specification with one property stating which location should 
be visited ($\varphi_1$) and another one for what should be done at visited locations 
($\varphi_2$). We also require $\varphi_0 = \G\F \hasNext $ to ensure that the plan 
continually 
processes locations from the iterator.
%
%

For property $\varphi_1$ we need to introduce three fluents:
$\FpatrolNextYes = \langle \yinP, \hasNext \rangle_{\emph{Initially}_{\bot}}$
that is true when the location selected by the iterator has been confirmed to be in the 
set of patrollable locations $P$ (\yinP). $\FrespondedPatrol = \langle 
\{\ninP,\yinP\}, 
\hasNext \rangle_{\emph{Initially}_{\bot}}$ is true when 
a response to \inP has been received. $\Farrived = \langle 
\arrived, \hasNext \rangle_{\emph{Initially}_{\bot}}$ is 
true when the UAV has arrived to the location selected by the iterator. 

The patrol condition  
 $\VisitCondition = \FrespondedPatrol \wedge  (\FpatrolNextYes \iff \Farrived )$, 
is that the \inP query has been 
responded, and the UAV has \arrived at that location if and only if the response was 
\yinP. Additionally,
$\varphi_1 =  \G(\yesNext \implies \neg \remove \W \VisitCondition)$
requires, for every new location that is selected by the iterator, to not remove that 
location from the iterator until the $\VisitCondition$ is achieved.

The specification of what to achieve at each visited location 
($\varphi_2$) follows a similar pattern. We use a fluent $\FYesFire = \langle 
\yFire, \hasNext \rangle_{\emph{Initially}_{\bot}}$ to model that fire has been 
detected at the current location, fluent $\FPhoto = \langle 
\photo,\hasNext \rangle_{\emph{Initially}_{\bot}}$ to model that a photo has been taken  
and $\FrespondedFire = \langle \{\yFire,\nFire\},\hasNext 
\rangle_{\emph{Initially}_{\bot}}$ to model reception of a response to \fire. 

The condition to be achieved once arrived at a location 
$ \ArrivedCondition = \FrespondedFire \wedge (\FYesFire  \iff \FPhoto)$
is that a response from the fire sensor must have been received and that a photo 
should be taken if and only if the response is positive. Consequently, we have
$\varphi_2 =  \G(\arrived \implies \neg \remove \W \ArrivedCondition)$.

If $E$, $\varphi_0$, $\varphi_1$, and $\varphi_2$ as defined above are fed to 
MTSA~\cite{MTSA} then the resulting controller is the one depicted in the top left 
of Figure~\ref{fig:architecture} (\emph{Iterator-Based Controller}).

%% file: Imagenes/LTS/iterator.tex
\centering
\TinyPicture
\VCDraw{
    \begin{VCPicture}{(-5.8,-0.6)(1.5,3.3)}
    \State[0]{(-4.2,2)}{0}
    \State[1]{(0,0)}{1}
    \State[2]{(0,2)}{2}
    \State[3]{(-4.2,0)}{3}
    \Initial[w]{0}

    \ChgEdgeLineStyle{dashed} 
    \EdgeL[.5]{0}{1}{\hasNext}
    \EdgeR{2}{0}{\remove}
    \EdgeR[.3]{3}{0}{\reset}
    \RstEdgeLineStyle 

    \EdgeR[.3]{1}{2}{\yesNext}
    \EdgeL{1}{3}{\noNext}
    \end{VCPicture}
}

%% file: Imagenes/LTS/SensorLTS.tex
\TinyPicture
\VCDraw{ %
    \begin{VCPicture}{(-1.5,-1.6)(2.8,2.3)}
    \State[0]{(-0.8,0)}{0}
    \State[1]{(2.2,0)}{1}
    \Initial[s]{0}

    \ChgEdgeLineStyle{dashed} 
    \EdgeL{0}{1}{\fire}
    \RstEdgeLineStyle 

\SetArcAngle{45} 
\SetLArcAngle{45} 

    \LArcR{1}{0}{\yFire}
    \LArcL{1}{0}{\nFire}
    \end{VCPicture}
}


%
%

%% file: Imagenes/LTS/SensorNextLTS.tex
\TinyPicture
\VCDraw{ %
    \begin{VCPicture}{(-1.5,-1.6)(3.8,2.3)}
    \State[0]{(-0.8,0)}{0}
    \State[1]{(3.2,0)}{1}
    \Initial[s]{0}

    \ChgEdgeLineStyle{dashed} 
        \EdgeL[.5]{0}{1}{\inP}
    \RstEdgeLineStyle 

\SetArcAngle{45} 
\SetLArcAngle{45} 

    \LArcR{1}{0}{\yinP}
    \LArcL{1}{0}{\ninP}

    \end{VCPicture}
}

%
%
%

%% file: Imagenes/LTS/constraintNext.tex
\TinyPicture
\VCDraw{ %
  \begin{VCPicture}{(-1,-1.6)(4.3,2.3)}
    \State[0]{(1,-1)}{0}
    \State[1]{(1,1)}{1}
    \Initial[w]{0}

    \ChgEdgeLineStyle{dashed} 
    \LArcL[.55]{1}{0}{\remove}
    \LoopE[.7]{0}{\remove}
    \LoopE[.2]{1}{\inP}
    \RstEdgeLineStyle 

    \LArcL{0}{1}{\yesNext}
    \end{VCPicture}
}

%% file: Imagenes/LTS/constraintCurrent.tex
\TinyPicture
\VCDraw{ %
  \begin{VCPicture}{(-1.2,-1.6)(3.8,2.3)}
    \State[0]{(1,-1)}{0}
    \State[1]{(1,1)}{1}
    \Initial[w]{0}

    \ChgEdgeLineStyle{dashed} 
    \LArcL[.55]{1}{0}{\hasNext}
    \LoopE[.7]{0}{\hasNext}
    \LoopE[.4]{1}{\fire}
     \LoopE[.2]{1}{\photo}
        \RstEdgeLineStyle 

    \LArcL{0}{1}{\arrived}
    \end{VCPicture}
}

%
%
%

%% file: Imagenes/LTS/capabilities.tex
\TinyPicture
\VCDraw{ %
  \begin{VCPicture}{(-1.2,-1.6)(3.5,2.3)}
    \State[0]{(1,-1)}{0}
    \State[1]{(1,1)}{1}
    \Initial[w]{0}
    \ChgEdgeLineStyle{dashed} 

    \LoopE[.7]{0}{\photo}
    \LoopE[.2]{1}{\photo}
      \LArcL{0}{1}{\go}
    
    \RstEdgeLineStyle 
    \LArcL[0.55]{1}{0}{\arrived}

    \end{VCPicture}
}

%% file: Imagenes/LTS/ConstratintGo.tex
 \TinyPicture
 \VCDraw{ %
  \begin{VCPicture}{(-0.5,-1.3)(6.5,2.6)}
     \State[1]{(1,-1)}{1}
    \State[0]{(3,1)}{0}
    \State[2]{(5,-1)}{2}
     \Initial[n]{0}

      \ChgEdgeLineStyle{dashed} 
     \EdgeR[.5]{1}{2}{\go}
    \EdgeL[.7]{1}{0}{\remove}
    \EdgeR[.7]{2}{0}{\remove}
    
  \RstEdgeLineStyle 
         \LArcL[.4]{0}{1}{\yesNext}
       
     \end{VCPicture}
 }

%% file: hybrid.tex
\section{HYBRID CONTROL LAYER} \label{sec:hybrid} 

A hybrid control layer  (e.g., \cite{Fainekos05,Gazit08})  provides an 
interface 
between a discrete controller and the lower level continuous 
control of the robot. Figure~\ref{fig:architecture} shows an  
architecture both for our iterator-based approach and one that manages 
locations explicitly at the discrete layer (e.g., 
\cite{ReactiveMissionMotion,Gazit08}).

In an \textit{iterator-based approach} the workspace is discretised 
independently of 
the 
synthesis procedure and fed to the Iterator module before the start of the 
mission. 
For the Fire Patrol task, the discretization also feeds the Patrollable Area 
Sensor with the locations that appear in orange  ($P = \set{\formatLoc{C5,A2,B2,\ldots 
}}$) in the 
map of 
Figure~\ref{fig:architecture}.

At runtime,  \hasNext, \yesNext and \noNext are used to loop over the 
discretised 
locations. In Figure~\ref{fig:architecture} the next location in the Iterator is 
$\formatLoc{C3}$. 
When the plan executes \inP, it produces a call to the Iterator (see red \caja{a} in 
Figure~\ref{fig:architecture}) that then forwards the request 
$\formatAction{Is.InP}(\formatLoc{C3})\formatAction{?}$ 
to the Patrollable Area sensor  \caja{b}. The sensor confirms that $\formatLoc{C3} \in P$
\caja{c} and the plan receives event \yesPatrolNext \caja{d}. 

Similarly, when \go is issued \caja{e}, the Iterator makes a
$\formatAction{go}(\formatLoc{C3})$ call to the motion planner \caja{f}. The motion 
planner 
generates the control inputs \caja{g}\caja{h} to reach $\formatLoc{C3}$ possibly also 
performing
static and dynamic obstacle avoidance.

Once the target location is reached, the \arrived event is propagated 
upwards \caja{i}\caja{j}  to the plan which then queries the existence of \fire 
\caja{k}. Note that as this query involves sensing the current robot location (and not 
 the Iterator's next location), the event  is sent directly through to the 
Fire 
Sensor.


In an \textit{explicit location approach}, synthesis requires information of the 
discretization to determine the order in which locations are to be visited. 
In the Fire Patrol example, the order in which locations in $P$ are to be patrolled is 
decided by the synthesis procedure (instead of the Iterator). 

Furthermore, the explicit location plan controls the path that must be followed by the 
robot, i.e. while the iterator-based approach set $\formatAction{C5}$ as the first 
patrol location to visit, the explicit-location plan sets the path to be followed to 
reach  $\formatLoc{C5}$: $\formatLoc{B7,  C7, C6, C5}$. This 
allows some static obstacle avoidance manoeuvres (e.g., 
\cite{Fainekos05,Wolff13ICRA}). 
 Nonetheless, the motion planner must still generate the control inputs between adjacent 
 discrete locations, deal with fine grained static obstacle avoidance and also  
 dynamic obstacle avoidance  (e.g., \cite{Vasile14}).
 
%
%
%

\textbf{Location Sorting:}
The order in which the Iterator offers locations to the discrete event controller can have 
significant impact.  Consider a Fire Patrol mission shown on the left side of 
Figure~\ref{fig:architecture}. The UAV started in location $\formatLoc{A7}$ and was 
offered $\formatLoc{B7}$ as the first location. As it is not an area to be patrolled, it 
was removed from 
the Iterator. This also occurred for $\formatLoc{C7}$, $\formatLoc{B6}$, and 
$\formatLoc{C6}$. Only when $\formatLoc{C5}$ was 
selected as 
the next element did the UAV \go to that location.
Many more locations that do not correspond to patrol areas could have been 
offered 
thus delaying the first \go command. In addition, a much more distant 
patrol location 
(e.g., $\formatLoc{G1}$) could have been selected, forcing possibly a less efficient 
patrol 
strategy. 

Consequently, an important component of the hybrid layer is the Sorter.
Our hybrid layer design works on the assumption that the best next location 
to offer is a function of the distance from the current robot location. This is a challenge as 
sorting must be done over a large set of locations regularly. Sorting is performed while the 
robot is travelling between the requested location (\go) and the moment it reaches it 
(\arrived). Distances are computed with respect  to the location that the robot will have 
once \arrived occurs. 

The sorting criteria must be simple 
enough to allow fast computing of each location's priority but not oversimplified, in 
order to produce acceptable overall task trajectories. In the next section we demonstrate 
experimentally that sorting over trajectory length using 
a simplified robot dynamic behaviour model allows fast enough sorting while 
providing reasonable trajectories for a fixed-wing UAV travelling at 
\SI{17}{\metre/\second}.

%% file: experiments.tex
\section{VALIDATION} \label{sec:experiments}
We  first show applicability to tasks taken 
from~\cite{ReactiveMissionMotion,Menghi19} 
and~\cite{mavproxy}, and we analyse scalability. Task specifications and results are 
available at~\cite{Repo}.

\subsection{Experimental Configuration}

All experiments were run on either a simulated or real fixed-wing 
battery-powered UAV. 

We used the robot in 
Figure~\ref{fig:architecture}, with low-level control provided by an 
off-the-shelf Pixhawk autopilot loaded with Ardupilot 
firmware \cite{Pixhawk} ArduPlane, and sensors providing information of 
the system's environment (e.g., Raspberry 
Pi Camera Module V2 for capturing ground images).


We built the discrete plan interpreter and hybrid control layer by extending the Ground 
Control Station (GCS) software 
MAVProxy~\cite{mavproxy} with custom Python modules. The hybrid 
control is run on an onboard Raspberry Pi 3B+ and communicates with 
the autopilot via the telemetry serial port.
We also used an instance of MAVProxy to allow human monitoring 
from the ground on a laptop which communicates with the 
autopilot via a SiK Telemetry Radio. Note, however, that mission execution 
is entirely run on onboard. 

For simulations, we replaced the plane and autopilot with 
the ArduPilot Software In The Loop (SITL) simulator that 
simulates the UAV's dynamics, autopilot and physical environment. This allows keeping the 
exact same onboard computer and hybrid control 
software as in the real flights. 
In the simulations we feature automatic takeoff and landing, while for 
safety reasons in the real flights a remote control 
(RC) radio system was connected to the Pixhawk to perform manual 
takeoff and landing.


The robot capability model used is an extension of Figure~\ref{fig:capabilities} to support  
taking-off and landing. Discrete event controllers were synthesised 
using MTSA~\cite{MTSA} and loaded onto the onboard computer before starting each 
mission.

\textbf{Motion Planning and Iterator Sorting.}
Although motion planning has been a greatly 
researched for both online and offline computation (e.g., 
\cite{Ji15,Bezzo16}), we implemented a fairly simple 
scheme, sufficient for our experimental goals, that does not consider 
static or dynamic obstacles.  

The motion planner generates sequences of control inputs for the 
autopilot based on trajectories that are computed by concatenating 
straight paths and turns for a given maximum turn radius, similar 
to~\cite{Hota13}. 
The planner finds (assuming constant speed and 
a maximum of two turns) 
 a trajectory for a 
given arrival direction at location  \Next  from the current  location and 
velocity vector of the UAV. 
We force the arrival direction to be parallel to 
a given fixed direction (e.g.,  grid axis to favour straight orderly grid coverage or 
perpendicular 
to the wind direction to increase flight stability).
 
The Sorter  uses the same trajectory computation. With a $\SI{50}{\metre} \times 
\SI{50}{\metre}$
discretization and a UAV that flies at 
\SI{17}{\metre/\second}, the minimum 
flight time between a \go command and an \arrived event is \SI{2.9}{\second}. In 
this 
time,  at least \num{40000} locations can be sorted on the onboard Raspberry Pi 
3B+. 
 
\subsection{Tasks}

\begin{figure}[bt]
\centering
  \vspace{1.5mm}
 \includegraphics[scale=0.23]{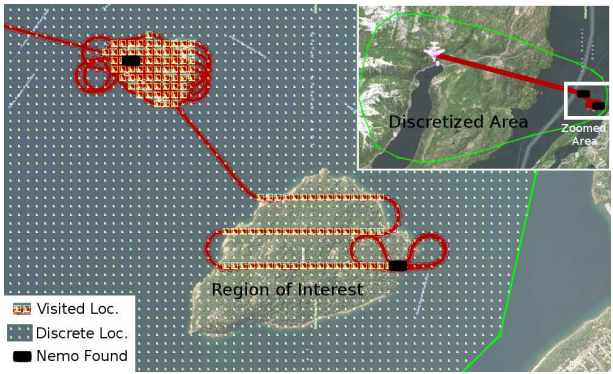}
\caption{Simulation of the Find Nemo task.}
\label{fig:nemo} 
\end{figure}

\subsubsection{Find Nemo}
The Find Nemo task~\cite{ReactiveMissionMotion} requires a robot with a Nemo sensor 
and camera to search for Nemo in \num{4} regions of interest out of a total of \num{12}. 
The task is to continuously, for all regions of interest, go, sense for Nemo, and if found 
stay and photograph.

We synthesised and ran a simulated task for \SI{437} regions of interest over a 
total of \SI{102307} regions. We used as locations of interest the two islands that can 
be seen in Figure~\ref{fig:nemo}. We randomised the appearance and disappearance of Nemo. 
Figure~\ref{fig:nemo} shows the UAV's path while searching and finding Nemo in two 
locations (one in each island), visiting this location until Nemo disappears and then 
resuming the search.

\begin{figure*}[bt]
\vspace{1.5mm}
\begin{subfigure}[h]{0.16\textwidth}
\centering
\vspace{3mm}
\includegraphics[scale=0.25]{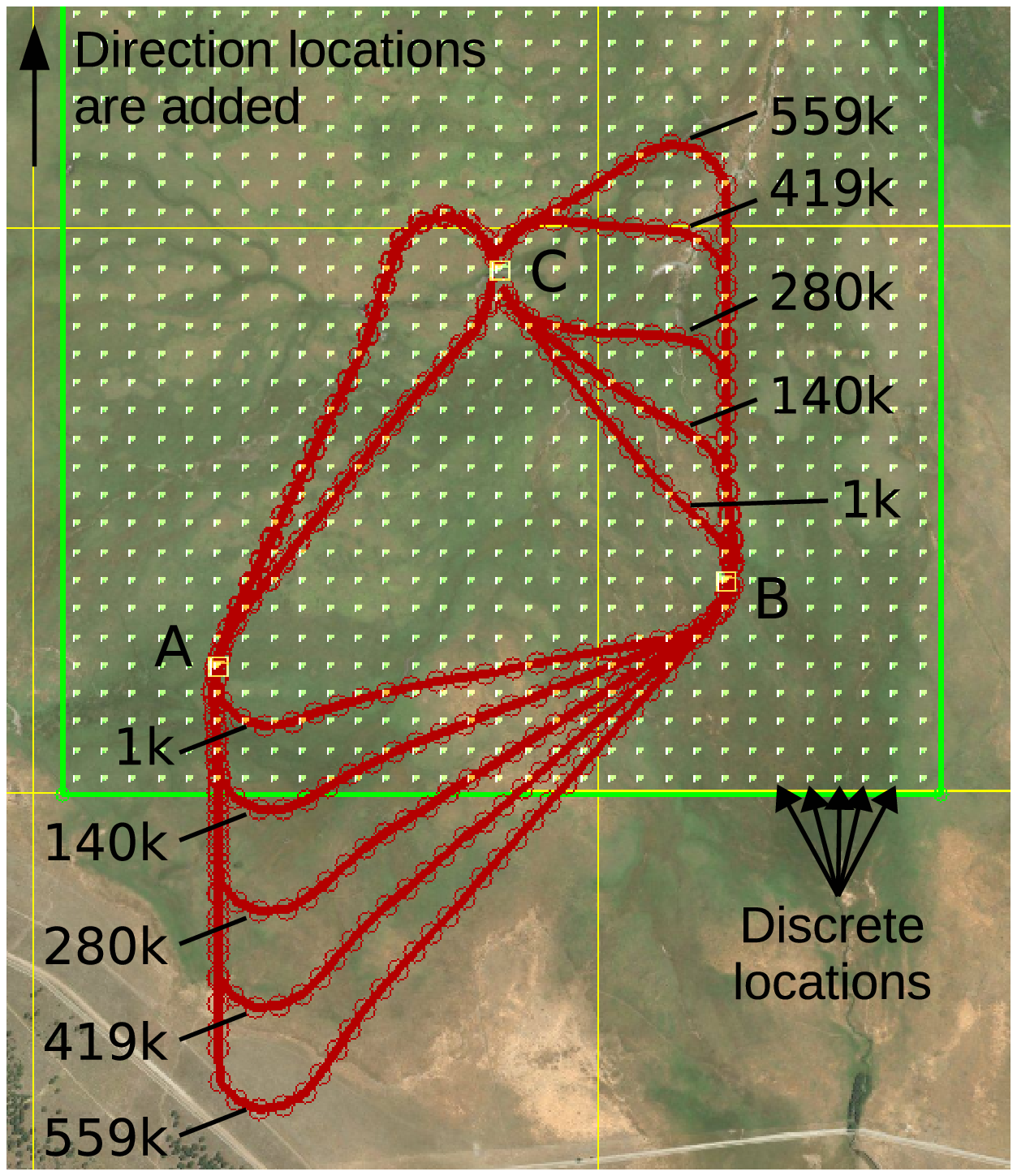}
\caption{Patrol trajectories}
\label{fig:patrolgraph}
\end{subfigure}
~
\begin{subfigure}[h]{0.252\textwidth}
\centering
\includegraphics[scale=0.217]{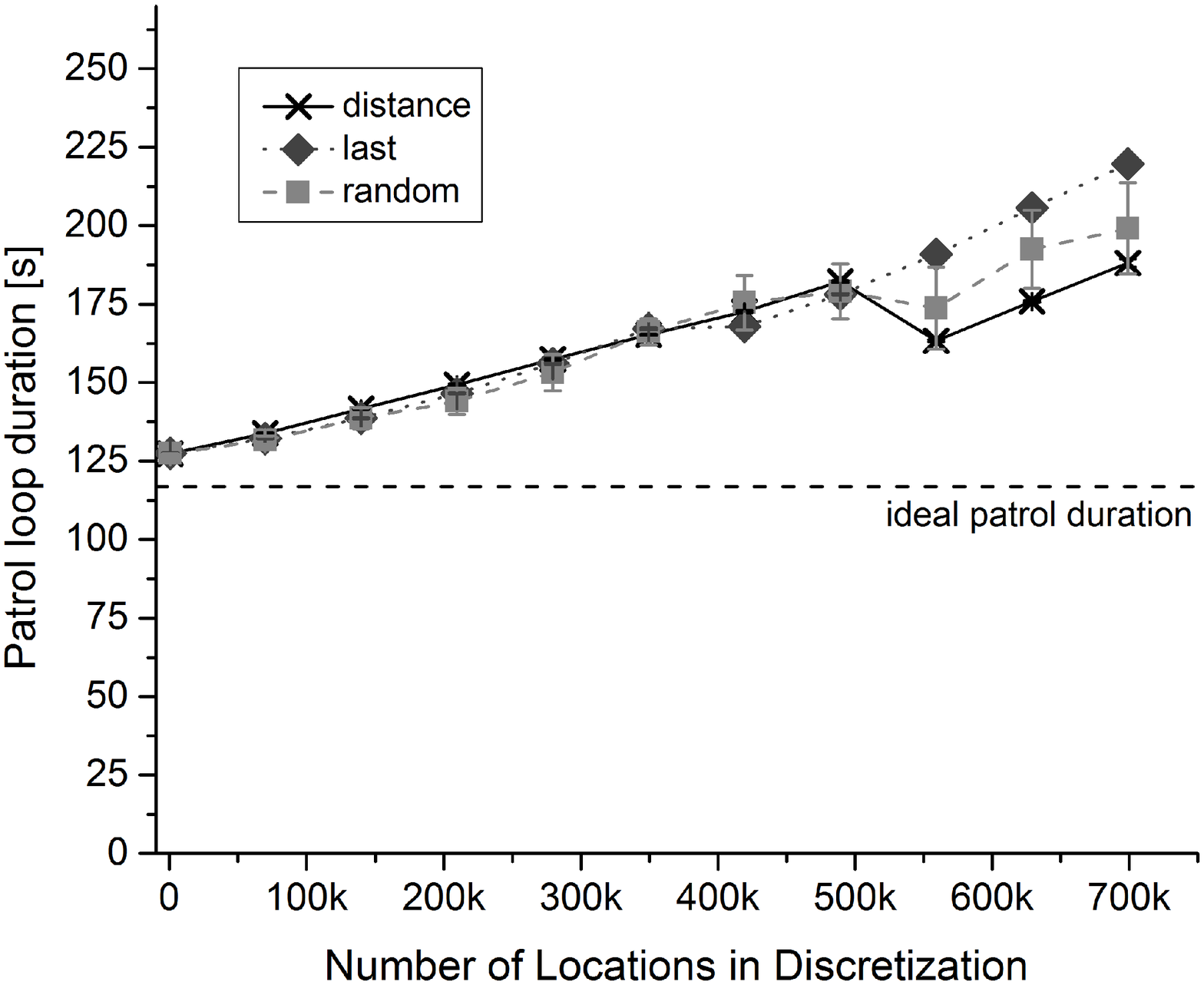}
\caption{Patrol loop time}
\label{fig:patrolloop}
\end{subfigure}
~
\begin{subfigure}[h]{0.245\textwidth}
\centering
\includegraphics[scale=0.213]{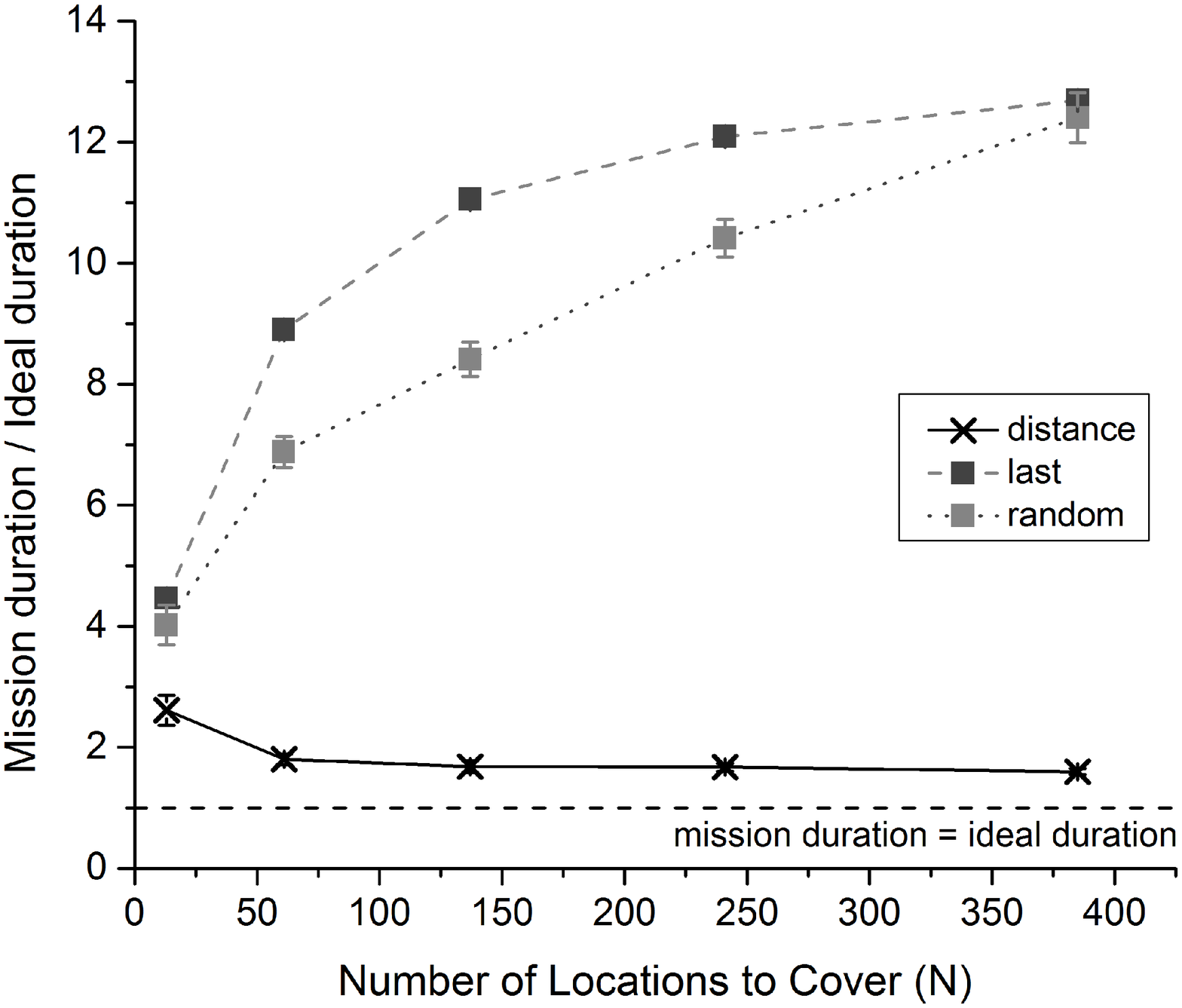}
\caption{Cover time of N locations}
\label{fig:coverN}
\end{subfigure}
~
\begin{subfigure}[h]{0.25\textwidth}
\centering
\includegraphics[scale=0.217]{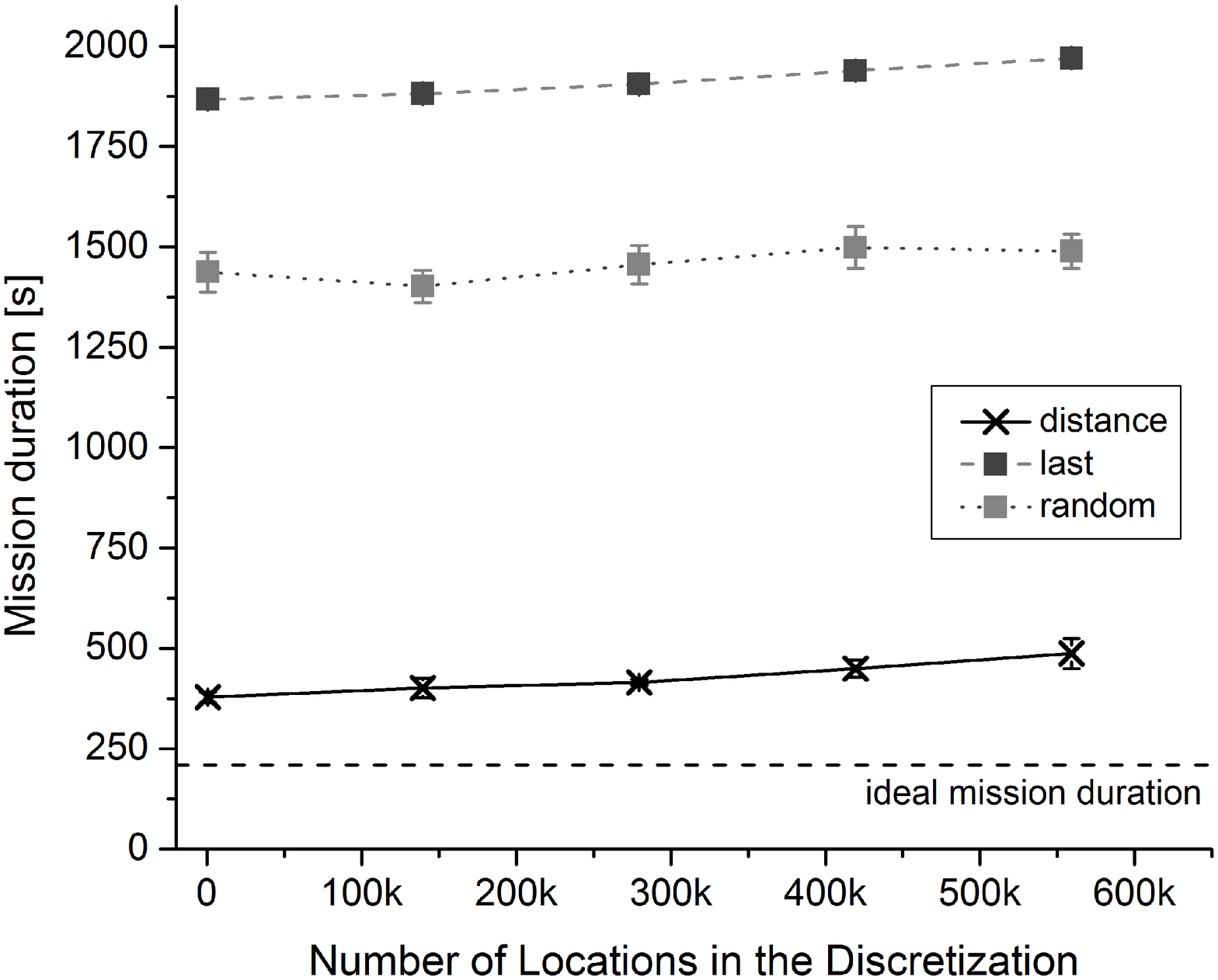}
\caption{Cover time vs disc. size}
\label{fig:cover61}
\end{subfigure}
\caption{(a) Patrol trajectories for different discretization 
sizes using \textit{last} sorter. (b) Loop duration for Ordered Patrol varying total 
number of locations and 
sorting criteria. (c) Proportional overhead for Cover task
varying  number of locations to cover, universe of locations fixed at \num{713}. (d) 
Duration 
of 
covering \num{61} locations varying total locations. 
\textit{Note:} The error bars are three times the estimated 
standard error of the mean, and can be smaller than the symbol.
Ideal duration is calculated using the minimum flight distance needed for the mission 
using constant speed, ignoring the UAV's movement restrictions.}
\label{fig:graphs}
\end{figure*}

\subsubsection{Search and Map Target}
Inspired on~\cite{mavproxy}, we specified and flew a task requiring to find a red 
target in a field and to map out all locations from which the target is visible and then land. 
The task specification can be structured as two modes. The first is a 
search, very much like Nemo, using a target sensor that captures an image and processes it 
to search for a red $\SI{2}{\metre} \times \SI{2}{\metre}$ target. If the target is 
found, the location is stored in a shared data structure. The second is a map mode in 
which all cells adjacent to one from which the target was seen must be visited, for which 
a second sensor responds \yesAdjacentNext if the location selected by the Iterator is 
adjacent to one in the shared data structure.

We ran our UAV for this task for \num{29502} locations. The estimated 
flight time of covering all locations optimally at $\SI{17}{\metre/\second}$ is over
\num{24} hours, significantly over the flight autonomy of the UAV (\SI{40}{\minute}  or 
$\sim$\num{800} locations). For this reason the red target 
was set relatively close to the launchpad. The video in~\cite{Repo} shows the flight 
of the UAV including both the trajectory as displayed by the monitoring ground station, 
the photos taken at locations and the mapped area. 



\subsubsection{Ordered Patrol}
\label{sec:orderedPatrol}
In~\cite{Menghi19}, a study of common UAV task requirements 
taken from over robotics papers is presented. One common requirement is that of an 
ordered 
patrol which 
requires visiting a set of locations in a particular order and at each one performing some 
task. In essence, this is the Fire Patrol task (which is another common 
pattern~\cite{Menghi19}, the unordered patrol), but with an order for which the patrol 
locations must be visited. This task requires explicit treatment of locations at 
the plan level. A sensor for each location is needed to include in the requirements the 
sequence in which these sensors must be used to find the places to visit.

We ran simulated versions of this task  which we report on in 
Section~\ref{sec:simulation} 
to analyse the scalability of our approach. 

\subsubsection{Cover}
\label{sec:cover}
Another requirement referred to in~\cite{Menghi19} is that of covering (i.e., visiting)
a statically defined region. The specification of such tasks in an 
iterator-based fashion does not differ significantly from the Fire Patrol task. The set of 
locations to be covered can be determined at runtime by the plan by using a sensor like 
\inP. We ran simulated 
versions of this task which we report on in Section~\ref{sec:simulation}.

\subsection{Synthesis Time Scalability}
The synthesis time for the tasks discussed in this 
paper was lower than \SI{5}{\second} on a laptop with a Intel i7 3.5GHz processor 
and 12GB of RAM. As discussed previously, this time is independent of the size of the 
locations over which the Iterator operates, which means that, except 
for Ordered Patrol, the number of locations can be scaled indefinitely. This includes 
both total locations and the locations to-patrol, of-interest, to-cover, and 
red locations in the Fire Patrol, Find Nemo, Cover, and Search and Map tasks.  
However, our approach is not independent of the number of sensors required to specify 
the task. Thus, for the Ordered Patrol
task, although the total number of locations can be scaled, the number of 
locations to be patrolled in a specific order cannot. Indeed, for $n$ locations to be 
orderly visited, the state space for synthesis will grow $2^n$ as in any 
synthesis approach~\cite{Wolff2013}. 

\subsection{Runtime Overhead} \label{sec:simulation}
To analyse the overhead (in terms of flight time) of iterating increasingly large location sets 
at runtime we selected, based on our understanding of our approach, best and worst case 
tasks.  To understand the impact of the Sorting component, we ran these tasks for three 
sorting strategies: the one described in Section~\ref{sec:hybrid} (distance), a highly 
inefficient one that puts the interesting elements to visit 
at the end of the iterator (last), and a random ordering (random). 

Our \textit{worst case task} is an Ordered Patrol  
(Section~\ref{sec:orderedPatrol}) because at runtime the Iterator may continuously 
offer last the next patrol location to be visited according to plan. This forces the plan 
to iterate over all locations every time before  \go. We simulated the Ordered Patrol 
of 3 locations as in~\cite{Wolff16} with variable amount of discrete locations.

Our \textit{best case task} is a Cover task as one iteration over the whole location 
universe suffices to complete the task. Of course, the order in which locations to be 
covered are offered may produce more or less efficient coverage paths.  
We simulated a cover task with variable amount of discrete locations and 
contiguous locations to be covered. 

For the Ordered Patrol task, we depict in Figure~\ref{fig:patrolloop} the time to visit all 
locations once for an increasing location universe size. Each data point is the average over 
\num{30} simulations for a particular sorting criteria. 
For the Cover task, we show in Figure~\ref{fig:coverN} how cover time increases as the 
number of locations to cover does. Mission duration (up to 
\num{3} hours each) required limiting experimentation to \num{6} samples per sorting 
criteria and 
size. Finally, we depict how the size of location universe impacts mission duration when 
requiring to cover \num{61} locations (Figure~\ref{fig:cover61}).

Figure~\ref{fig:patrolloop} shows that degradation seems linear 
in the number of total locations and that the three sorting strategies make little difference 
in relative terms when comparing the overall mission duration to the ideal mission 
duration. The overhead for \num{700000} locations is at most 
\SI{88}{\percent} over an ideal patrol mission. Figure~\ref{fig:patrolgraph} 
exemplifies why the the 
UAV's trajectory between the three locations to patrol degrades. As the universe of 
locations increases, so does the distance the UAV is flying in a straight line while 
iterating over all these locations to get the next patrollable location.  



In contrast,  Figure~\ref{fig:coverN} shows that sorting based on trajectory distance makes 
a big difference and provides near constant proportional overhead 
($\SI{60}{\percent}$) when increasing the size 
of region to cover. Figure~\ref{fig:cover61} also shows relevance of the sorting criteria and 
near constant duration covering a \num{61} location size while increasing the universe of 
locations. 


 

%% file: conclusions.tex
\section{Related Work}
The state explosion problem in temporal mission planning has been addressed in various 
ways.
$i)$ Improved online/offline motion planning 
(e.g.,~\cite{Livingston13,Cabreira19,Yan17,Ghosh16}) is orthogonal to our approach 
and 
can be introduced within our hybrid layer replacing the Sorter  and Motion Planner 
components. We do not compare empirically against these approaches, rather we  
show that even simple sorting and motion strategies already yield reasonable 
results. 
$ii)$ 
Advances in synthesis efficiency 
(e.g.,~\cite{TuLiP,LTLMoP,EfficientSynthesis}) are also orthogonal. An iterator 
abstraction can be used with a variety of approaches  to synthesis to scale task 
planning orders of magnitude beyond what can be 
achieved when all  universally quantified locations must be explicitly referred to.
$iii)$ Alternative strategies for integrating task and motion planning. The distinctive 
feature of the strategy presented in this paper is the use of an Iterator coupled 
with runtime 
motion planning. In \cite{Nedunuri14} a plan outline is manually constructed 
without explicit naming of locations, later to be filled offline by an SMT solver. 
Scale is limited by the solver, which depends on the number of locations. Indeed, as in 
another SMT-based approach~\cite{Kundu19} reported cases are below \num{800} discrete 
locations.
In~\cite{Wolff2013}  robot's paths are produced for LTL specifications by 
solving constrained reachability problems, but total complexity depends on the 
number 
of locations.
\cite{Fainekos05,Castro15} combine task and motion requirements in a GR(1) 
specification. Complexity of GR(1) is polynomial respect to the state space which 
grows combinatorially to the number of locations. \cite{Kaelbling11} propose a 
highly hierarchical approach which is able to solve 
large complex workspaces, but requires domain-dependent choices for the 
hierarchy. 

In all these approaches, increasing the 
number of universally quantified locations beyond a couple of thousand implies not 
being able to compute a plan. In contrast, in our approach, the increase in locations 
does not impede producing a plan, albeit with degraded mission trajectories 
due to runtime motion planning.


\section{CONCLUSIONS}


We propose iterator-based task planning to provide constant synthesis complexity 
with respect to the number of discrete locations universally quantified in a task 
specification. Iterator-based plans run on a hybrid control layer that performs 
runtime motion planning.  We show that simple location prioritisation and motion 
planning strategies suffice to provide adequate 
mission behaviour for iterator-based plans both in simulated and real UAV missions.

%% file: paper.bbl
\begin{thebibliography}{10}
\providecommand{\url}[1]{#1}
\csname url@rmstyle\endcsname
\providecommand{\newblock}{\relax}
\providecommand{\bibinfo}[2]{#2}
\providecommand\BIBentrySTDinterwordspacing{\spaceskip=0pt\relax}
\providecommand\BIBentryALTinterwordstretchfactor{4}
\providecommand\BIBentryALTinterwordspacing{\spaceskip=\fontdimen2\font plus
\BIBentryALTinterwordstretchfactor\fontdimen3\font minus
  \fontdimen4\font\relax}
\providecommand\BIBforeignlanguage[2]{{%
\expandafter\ifx\csname l@#1\endcsname\relax
\typeout{** WARNING: IEEEtran.bst: No hyphenation pattern has been}%
\typeout{** loaded for the language `#1'. Using the pattern for}%
\typeout{** the default language instead.}%
\else
\language=\csname l@#1\endcsname
\fi
#2}}

\bibitem{Fainekos05}
G.~E. Fainekos, H.~Kress-Gazit, and G.~J. Pappas, ``Temporal logic motion
  planning for mobile robots,'' in \emph{Proceedings of the 2005 IEEE
  International Conference on Robotics and Automation}, 2005, pp. 2020--2025.

\bibitem{Wolff13ICRA}
E.~M. {Wolff}, U.~{Topcu}, and R.~M. {Murray}, ``Efficient reactive controller
  synthesis for a fragment of linear temporal logic,'' in \emph{2013 IEEE
  International Conference on Robotics and Automation}, May 2013, pp.
  5033--5040.

\bibitem{Castro15}
J.~A. DeCastro, V.~Raman, and H.~Kress-Gazit, ``Dynamics-driven adaptive
  abstraction for reactive high-level mission and motion planning,'' in
  \emph{2015 IEEE International Conference on Robotics and Automation (ICRA)},
  2015, pp. 369--376.

\bibitem{piterman06}
N.~Piterman, A.~Pnueli, and Y.~Sa'ar, ``{Synthesis of reactive (1) designs},''
  \emph{Lecture notes in computer science}, vol. 3855, pp. 364--380, 2006.

\bibitem{Mapping}
A.~{Tariq}, S.~M. {Osama}, and A.~{Gillani}, ``Development of a low cost and
  light weight uav for photogrammetry and precision land mapping using aerial
  imagery,'' in \emph{2016 International Conference on Frontiers of Information
  Technology (FIT)}, 2016, pp. 360--364.

\bibitem{ReactiveMissionMotion}
H.~Kress-Gazit, G.~E. Fainekos, and G.~J. Pappas, ``Temporal-logic-based
  reactive mission and motion planning,'' \emph{IEEE Transactions on Robotics},
  vol.~25, no.~6, pp. 1370--1381, 2009.

\bibitem{Wolff2013}
E.~M. Wolff, U.~Topcu, and R.~M. Murray, ``Automaton-guided controller
  synthesis for nonlinear systems with temporal logic,'' in \emph{2013 IEEE/RSJ
  International Conference on Intelligent Robots and Systems}, 2013, pp.
  4332--4339.

\bibitem{Waldo}
H.~Kress-Gazit, G.~E. Fainekos, and G.~J. Pappas, ``Where's waldo? sensor-based
  temporal logic motion planning,'' in \emph{Proceedings 2007 IEEE
  International Conference on Robotics and Automation}, 2007, pp. 3116--3121.

\bibitem{Livingston13}
S.~C. Livingston and R.~M. Murray, ``Just-in-time synthesis for reactive motion
  planning with temporal logic,'' in \emph{2013 IEEE International Conference
  on Robotics and Automation}, 2013, pp. 5048--5053.

\bibitem{Nedunuri14}
S.~{Nedunuri}, S.~{Prabhu}, M.~{Moll}, S.~{Chaudhuri}, and L.~E. {Kavraki},
  ``Smt-based synthesis of integrated task and motion plans from plan
  outlines,'' in \emph{2014 IEEE International Conference on Robotics and
  Automation (ICRA)}, May 2014, pp. 655--662.

\bibitem{Vasile14}
C.~I. Vasile and C.~Belta, ``Reactive sampling-based temporal logic path
  planning,'' in \emph{2014 IEEE International Conference on Robotics and
  Automation (ICRA)}, 2014, pp. 4310--4315.

\bibitem{Menghi19}
C.~{Menghi}, C.~{Tsigkanos}, T.~{Berger}, and P.~{Pelliccione}, ``Psalm:
  Specification of dependable robotic missions,'' in \emph{2019 IEEE/ACM 41st
  International Conference on Software Engineering: Companion Proceedings
  (ICSE-Companion)}, May 2019, pp. 99--102.

\bibitem{Gazit08}
H.~Kress-Gazit, G.~Fainekos, and G.~Pappas, ``Translating structured english to
  robot controllers,'' \emph{Advanced Robotics}, vol.~22, pp. 1343--1359, 10
  2008.

\bibitem{Ji15}
X.~{Ji} and J.~{Li}, ``Online motion planning for uav under uncertain
  environment,'' in \emph{2015 8th International Symposium on Computational
  Intelligence and Design (ISCID)}, vol.~2, Dec 2015, pp. 514--517.

\bibitem{Bezzo16}
N.~{Bezzo}, K.~{Mohta}, C.~{Nowzari}, I.~{Lee}, V.~{Kumar}, and G.~{Pappas},
  ``Online planning for energy-efficient and disturbance-aware uav
  operations,'' in \emph{2016 IEEE/RSJ International Conference on Intelligent
  Robots and Systems (IROS)}, Oct 2016, pp. 5027--5033.

\bibitem{mavproxy}
H.~{Choi}, M.~{Geeves}, B.~{Alsalam}, and F.~{Gonzalez}, ``Open source
  computer-vision based guidance system for uavs on-board decision making,'' in
  \emph{2016 IEEE Aerospace Conference}, 2016, pp. 1--5.

\bibitem{Wei18}
M.~{Wei} and V.~{Isler}, ``Coverage path planning under the energy
  constraint,'' in \emph{2018 IEEE International Conference on Robotics and
  Automation (ICRA)}, May 2018, pp. 368--373.

\bibitem{Keller76}
R.~Keller, ``Formal verification of parallel programs.'' \emph{Communications
  of the ACM}, vol.~19, pp. 371--384, 07 1976.

\bibitem{gianna03}
D.~Giannakopoulou and J.~Magee, ``Fluent model checking for event-based
  systems,'' in \emph{Proceedings of the 9th European Software Engineering
  Conference Held Jointly with 11th ACM SIGSOFT International Symposium on
  Foundations of Software Engineering}, ser. ESEC/FSE-11.\hskip 1em plus 0.5em
  minus 0.4em\relax New York, NY, USA: ACM, 2003, pp. 257--266.

\bibitem{MTSA}
N.~R. D'Ippolito, V.~Braberman, N.~Piterman, and S.~Uchitel, ``Synthesis of
  live behaviour models,'' in \emph{Proceedings of the eighteenth ACM SIGSOFT
  international symposium on Foundations of software engineering}, ser. FSE
  '10.\hskip 1em plus 0.5em minus 0.4em\relax New York, NY, USA: ACM, 2010, pp.
  77--86.

\bibitem{Lamport77}
L.~Lamport, ``Proving the correctness of multiprocess programs,'' \emph{IEEE
  Transactions on Software Engineering}, vol.~3, no.~2, pp. 125--143, Mar.
  1977.

\bibitem{Kundu19}
T.~{Kundu} and I.~{Saha}, ``Energy-aware temporal logic motion planning for
  mobile robots,'' in \emph{2019 International Conference on Robotics and
  Automation (ICRA)}, May 2019, pp. 8599--8605.

\bibitem{StateOfTheArt}
C.~Belta, A.~Bicchi, M.~Egerstedt, E.~Frazzoli, E.~Klavins, and G.~J. Pappas,
  ``Symbolic planning and control of robot motion [grand challenges of
  robotics],'' \emph{IEEE Robotics Automation Magazine}, vol.~14, no.~1, pp.
  61--70, 2007.

\bibitem{ControlModes}
A.~Balkan, M.~Vardi, and P.~Tabuada, ``Mode-target games: Reactive synthesis
  for control applications,'' \emph{IEEE Transactions on Automatic Control},
  vol.~63, no.~1, pp. 196--202, 2018.

\bibitem{Repo}
\BIBentryALTinterwordspacing
Repository Package, accessed 2019-09-15. [Online]. Available:
  \url{https://bitbucket.org/szudaire/iterator-based-planning}
\BIBentrySTDinterwordspacing

\bibitem{Pixhawk}
A.~Sinisterra, M.~Dhanak, and N.~Kouvaras, ``A usv platform for surface
  autonomy,'' in \emph{OCEANS 2017 - Anchorage}, 2017, pp. 1--8.

\bibitem{Hota13}
S.~Hota and D.~Ghose, ``Optimal transition trajectory for waypoint following,''
  in \emph{2013 IEEE International Conference on Control Applications (CCA)},
  2013, pp. 1030--1035.

\bibitem{Wolff16}
E.~M. Wolff and R.~M. Murray, \emph{Optimal Control of Nonlinear Systems with
  Temporal Logic Specifications}.\hskip 1em plus 0.5em minus 0.4em\relax
  Springer International Publishing, 2016, pp. 21--37.

\bibitem{Cabreira19}
T.~M.~Cabreira, L.~Brisolara, and P.~Ferreira~Jr, ``Survey on coverage path
  planning with unmanned aerial vehicles,'' \emph{Drones}, vol.~3, p.~4, 01
  2019.

\bibitem{Yan17}
Z.~{Yan}, J.~{He}, and J.~{Li}, ``An improved multi-auv patrol path planning
  method,'' in \emph{2017 IEEE International Conference on Mechatronics and
  Automation (ICMA)}, Aug 2017, pp. 1930--1936.

\bibitem{Ghosh16}
M.~{Ghosh}, S.~{Thomas}, M.~{Morales}, S.~{Rodriguez}, and N.~M. {Amato},
  ``Motion planning using hierarchical aggregation of workspace obstacles,'' in
  \emph{2016 IEEE/RSJ International Conference on Intelligent Robots and
  Systems (IROS)}, Oct 2016, pp. 5716--5721.

\bibitem{TuLiP}
T.~Wongpiromsarn, U.~Topcu, N.~Ozay, H.~Xu, and R.~M. Murray, ``Tulip: A
  software toolbox for receding horizon temporal logic planning,'' in
  \emph{Proceedings of the 14th International Conference on Hybrid Systems:
  Computation and Control}, ser. HSCC '11.\hskip 1em plus 0.5em minus
  0.4em\relax New York, NY, USA: ACM, 2011, pp. 313--314.

\bibitem{LTLMoP}
C.~{Finucane}, {Gangyuan Jing}, and H.~{Kress-Gazit}, ``Ltlmop: Experimenting
  with language, temporal logic and robot control,'' in \emph{2010 IEEE/RSJ
  International Conference on Intelligent Robots and Systems}, Oct 2010, pp.
  1988--1993.

\bibitem{EfficientSynthesis}
S.~Dathathri and R.~M. Murray, ``Decomposing gr(1) games with singleton
  liveness guarantees for efficient synthesis,'' in \emph{2017 IEEE 56th Annual
  Conference on Decision and Control (CDC)}, 2017, pp. 911--917.

\bibitem{Kaelbling11}
L.~P. {Kaelbling} and T.~{Lozano-Pérez}, ``Hierarchical task and motion
  planning in the now,'' in \emph{2011 IEEE International Conference on
  Robotics and Automation}, May 2011, pp. 1470--1477.

\end{thebibliography}
